\documentclass[11pt]{article}

\usepackage[final]{acl}

\usepackage{times}
\usepackage{latexsym}

\usepackage[T1]{fontenc}

\usepackage[utf8]{inputenc}

\usepackage{microtype}

\usepackage{inconsolata}

\usepackage{graphicx}

\usepackage{graphicx} 
\usepackage{enumitem}
\usepackage{subcaption}
\usepackage{booktabs}
\usepackage{amsfonts}
\usepackage{amsmath}
\usepackage{booktabs}
\usepackage{multirow}
\usepackage{graphicx}
\usepackage{enumitem}
\usepackage{textcomp}
\usepackage{booktabs}

\usepackage{array}
\usepackage{bm}
\usepackage{tabularx}
\usepackage{tabulary}
\usepackage{array}
\usepackage{color,ulem}
\usepackage{xcolor}
\usepackage{caption}
\usepackage{makecell} 
\usepackage{xcolor} 
\usepackage{fontawesome5} 
\usepackage{hyperref}
\usepackage[table]{xcolor}

\usepackage{algorithm}
\usepackage{algorithmic}
\usepackage{array}
\usepackage{bm}
\usepackage{array}
\usepackage{float}  
\usepackage{placeins}

\title{CE-GPPO: Coordinating Entropy via Gradient-Preserving Clipping Policy Optimization in Reinforcement Learning}

\author{
Zhenpeng Su\textsuperscript{\rm 1}\footnotemark[1] \quad Leiyu Pan\textsuperscript{\rm 1}\footnotemark[1]  \quad Minxuan Lv\textsuperscript{\rm 1}\quad Yuntao Li\textsuperscript{\rm 2} \quad  \textbf{Wenping Hu}\textsuperscript{\rm 1} \\ \textbf{Fuzheng Zhang}\textsuperscript{\rm 1} \quad \textbf{Kun Gai}\textsuperscript{\rm 1} \quad \textbf{Guorui Zhou}\textsuperscript{\rm 1} \footnotemark[2] \\  
  \textsuperscript{\rm 1}Kuaishou Technology
  \textsuperscript{\rm 2}Independent\\
  \faEnvelope\ \href{suzhenpeng13@163.com}{suzhenpeng13@163.com}
}

\begin{document}
\maketitle
\renewcommand{\thefootnote}{\fnsymbol{footnote}} 
\footnotetext[1]{Equal contribution. This work was completed by Leiyu Pan during an internship at Kuaishou.} \footnotetext[2]{Corresponding authors. } 
\renewcommand{\thefootnote}{\arabic{footnote}}
\begin{abstract}
Reinforcement learning (RL) has become a powerful paradigm for optimizing large language models (LLMs) to handle complex reasoning tasks. A core challenge in this process lies in managing policy entropy, which reflects the balance between exploration and exploitation during training. Existing methods, such as proximal policy optimization (PPO) and its variants, discard valuable gradient signals from low-probability tokens due to the clipping mechanism. We systematically analyze the entropy dynamics and reveal that these clipped tokens play a critical yet overlooked role in regulating entropy evolution.
We propose \textbf{C}oordinating \textbf{E}ntropy via \textbf{G}radient-\textbf{P}reserving \textbf{P}olicy \textbf{O}ptimization (CE-GPPO), a novel algorithm that reintroduces gradients from clipped tokens in native PPO in a gentle and bounded manner. By controlling the magnitude of gradients from tokens outside the clipping interval, CE-GPPO is able to achieve an exploration-exploitation trade-off. We provide theoretical justification and empirical evidence showing that CE-GPPO effectively mitigates entropy instability.
Extensive experiments on mathematical reasoning benchmarks show that CE-GPPO consistently outperforms strong baselines across different model scales. 
\end{abstract}

\section{Introduction}

Reinforcement learning (RL) has increasingly become a paradigm for fine-tuning large language models (LLMs), shifting the focus from mere imitation learning to goal-directed optimization \citep{DBLP:conf/nips/Ouyang0JAWMZASR22, DBLP:journals/corr/abs-2402-03300, DBLP:journals/corr/abs-2501-12948}. Unlike supervised learning, which only fits the observed data distribution, RL directly optimizes model behavior through environmental feedback, enabling improvements on abstract objectives such as factual accuracy, coherence, and reasoning capability \citep{DBLP:journals/corr/abs-2503-14476}. In particular, Reinforcement Learning with Verifiable Rewards (RLVR) have attracted growing attention \citep{DBLP:journals/corr/abs-2411-15124}. By providing regularized and automatically evaluable reward signals, these approaches offer stable and interpretable guidance, significantly enhancing model performance in reasoning tasks \citep{su2025klear}.

Despite the promise of RL for goal-driven model optimization, training dynamics remain challenging, particularly in regulating policy entropy, a key indicator of the model’s exploration capability \citep{DBLP:conf/icml/AhmedR0S19}. Policy entropy measures the uncertainty in action selection \citep{DBLP:journals/corr/abs-2506-14758}.
Our in-depth analysis reveals that the dynamic behavior of entropy in RL can be traced to the intrinsic interaction between the advantage function and the token probability distribution. Specifically, gradient updates can be categorized into four typical patterns:

\begin{itemize}[leftmargin=*]
    \item Positive-advantage high-probability (PA\&HP) tokens and negative-advantage low-probability (NA\&LP) tokens: optimizing these tokens reinforces high-probability choices, accelerating policy convergence and entropy collapse.
    \item Positive-advantage low-probability 
    (PA\&LP) tokens and negative-advantage high-probability (NA\&HP) tokens: optimizing these tokens encourages exploration of low-probability actions, maintaining response diversity and mitigating entropy collapse.
\end{itemize}

\begin{figure*}[t]
    \centering
    \includegraphics[width=0.94\textwidth]{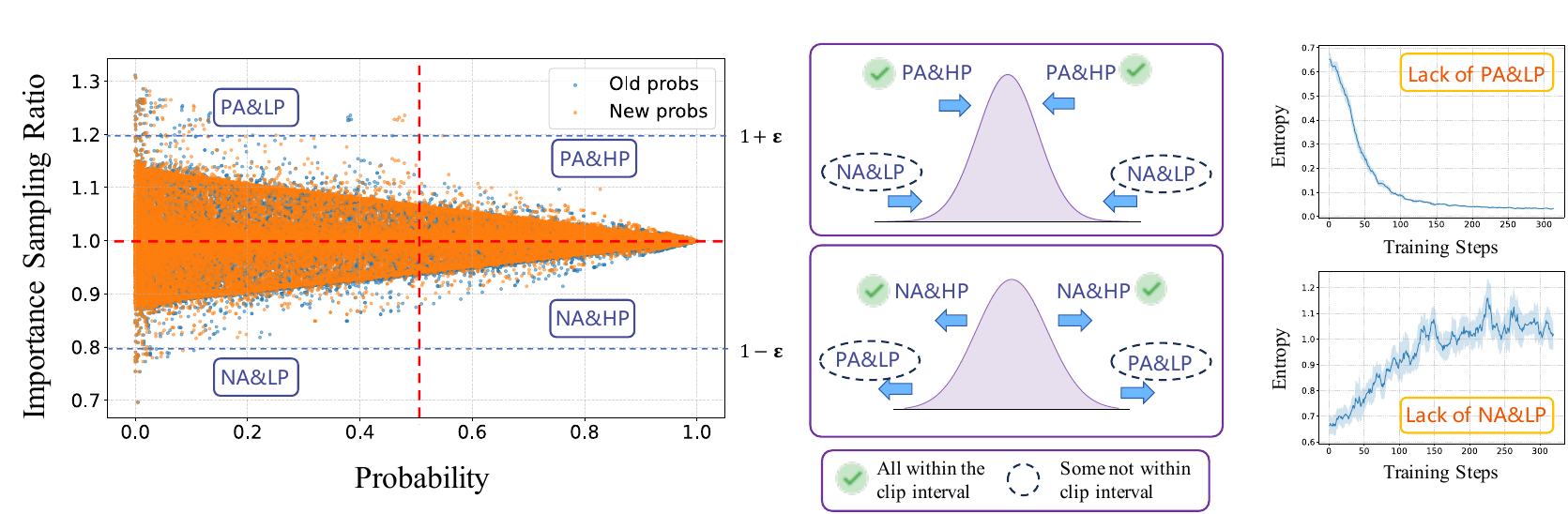}
    \caption{\textit{Left}: Importance sampling distribution of tokens with different probabilities. Based on the distribution, all tokens can be categorized into four types: PA\&HP, NA\&LP, PA\&LP and NA\&HP. \textit{Center}: The effect of the four token types on entropy dynamics. The two categories shown at the top contribute to entropy reduction, while those at the bottom contribute to entropy increase. Green check marks indicate tokens that lie within the clipping interval, whereas dashed circles denote tokens that partly fall outside the clipping interval. \textit{Right}: Entropy instability curves caused by the absence of some PA\&LP or NA\&LP tokens.}
    \label{fig:prob_ratio}
\end{figure*}

Further investigation revealed a strong connection between token probabilities and importance sampling. In preliminary experiments, we conducted RL training on the DeepSeek-R1-Distill-Qwen-7B model using datasets from the mathematics and code reasoning domains, and consistently observed the token distribution pattern illustrated in Figure~\ref{fig:prob_ratio}. Specifically, high-probability tokens typically lay within the PPO clipping interval~\citep{DBLP:journals/corr/SchulmanWDRK17}, whereas tokens outside the clipping interval were predominantly low-probability tokens.
For RL, clipped importance sampling is a commonly used technique. The clipping mechanism primarily controls the magnitude of updates to the policy model to ensure training stability. This results in a primary focus on optimizing unclipped high-probability tokens, but we find that ignoring the clipped low-probability tokens, i.e., NA\&LP and PA\&LP tokens, introduces following issues:

\begin{itemize}[leftmargin=*]
\item \textbf{Entropy collapse due to the absence of PA\&LP tokens.} Tokens truncated beyond the importance sampling threshold $1+\epsilon$, namely PA\&LP tokens, often include high-entropy tokens that correspond to valuable exploratory behaviors at critical decision points. Directly clipping the gradients of these tokens restricts exploration, leading to entropy collapse. Although DAPO mitigates this issue by extending the upper clipping bound to $1+\epsilon_h$ through the clip-higher strategy, high-entropy tokens that exceed this bound still suffer from the same problem.
\item \textbf{Entropy explosion due to the absence of NA\&LP tokens.} Tokens truncated beyond the importance sampling threshold $1-\epsilon$, namely NA\&LP tokens, include tokens that guide the model toward exploitation. Clipping the gradients of these tokens forces the model into excessive exploration, delays convergence, and consequently induces entropy explosion.
\end{itemize}

A natural idea is to merge the gradients of tokens outside the clipping range: PA\&LP tokens promote exploration, while NA\&LP tokens encourage exploitation. By respectively leveraging their gradient magnitudes to different extents, it is possible to coordinate policy entropy and strike a balance between exploration and exploitation, thus ensure entropy stability during training.

Based on these insights, we propose \textbf{CE-GPPO}, which reframes the control of entropy dynamics as managing gradients from tokens outside the clipping interval. 
Specifically, CE-GPPO uses a stop-gradient operation to include gradients from tokens beyond the clipping interval and adjusts their magnitude to maintain policy entropy at a high and stable level, which we find to be beneficial for improving model performance.
Importantly, we provide both theoretical and empirical evidence showing that incorporating gradients from tokens outside the PPO trust region, i.e., the tokens clipped by PPO, does not cause the policy model to deviate excessively from the old policy model and still preserves stable training.
Additionally, we observe that assigning \textit{greater weight to the gradients of PA\&LP tokens while less weight to those of NA\&LP tokens} helps the model maintain its exploration capability and achieve better performance than other strong baseline models.
The main contributions of our work can be summarized as follows:

\begin{itemize}[leftmargin=*]
    \item We reveal the intrinsic mechanism of entropy dynamics in RL for LLMs and identify \textbf{a novel perspective for controlling entropy evolution}.
    \item We propose CE-GPPO, an algorithm that regulates gradients from tokens outside both sides of the clip interval, enabling \textbf{ fine-grained control of policy entropy and update stability}.
    \item We empirically show that CE-GPPO achieves \textbf{effective coordination of policy entropy}, stabilizing the exploration–exploitation trade-off, and exhibiting strong hyperparameter robustness.
\end{itemize}

\section{Preliminary}

\subsection{Policy Optimization Algorithms}

\paragraph{Proximal Policy Optimization (PPO)} PPO \citep{DBLP:journals/corr/SchulmanWDRK17} is a widely used policy gradient method in RL, designed to balance learning stability and sample efficiency. It improves upon classical policy gradient approaches by constraining the magnitude of policy updates,  preventing destructive updates that could destabilize training. Concretely, its objective function is as follows:

\begin{small}
\begin{equation}
\begin{aligned}
\mathcal{J}_{\text{PPO}}(\theta)=\mathbb{E}_{x \sim \mathcal{D}, y \sim \pi_{\theta_{\text{old}}}(\cdot \mid x)}\left[\frac{1}{|y|} \sum_{t=1}^{|y|} \min \left(r_{t}(\theta) \hat{A}_{t}, \right.\right.\\
\left.\left.\text{clip}\left(r_{t}(\theta), 1-\epsilon, 1+\epsilon\right) \hat{A}_{t}\right)\right]
\label{ppo_loss}
\end{aligned}
\end{equation}
\end{small}
Here, $x$ denotes a prompt sampled from the data distribution $\mathcal{D}$, and $y = (y_1, \dots, y_{|y|})$ are output sequences sampled from the old policy $\pi_{\theta_\text{old}}$. The term $r_t(\theta) = \frac{\pi_\theta(y_t \mid x, y_{<t})}{\pi_{\theta_\text{old}}(y_t \mid x, y_{<t})}$ is the importance sampling ratio. $\hat{A}_t$ represents the estimated advantage, often computed using Generalized Advantage Estimation (GAE) \citep{DBLP:journals/corr/SchulmanMLJA15}. $\epsilon$ is a hyperparameter controlling the clipping range.


\paragraph{Group Relative Policy Optimization (GRPO)} \citet{DBLP:journals/corr/abs-2402-03300} introduces a critic-free RL method GRPO that simplifies policy optimization by eliminating explicit value function estimation. For each prompt $x$, it estimates advantages by normalizing the rewards among a group of $G$ sampled responses $\{r_i\}_{i=1}^G$.

\begin{small}
\begin{equation}
\hat{A}_{i,t} = \frac{r_i - \text{mean}(\{r_i\}_{i=1}^G)}{\text{std}(\{r_i\}_{i=1}^G)}
\end{equation}
\end{small}
The GRPO objective integrates this advantage estimation into a clipped policy gradient framework:

\begin{small}
\begin{equation}
\begin{aligned}
\mathcal{J}_{\text{GRPO}}(\theta) = \mathbb{E} \left[ \frac{1}{G} \sum_{i=1}^{G} \frac{1}{|y_i|} \sum_{t=1}^{|y_i|} \min \big( r_{i,t}(\theta) \hat{A}_{i,t}, \right.\\
\left. \text{clip}(r_{i,t}(\theta), 1-\epsilon, 1+\epsilon) \hat{A}_{i,t} \big) \right]
\end{aligned}
\end{equation}
\end{small}

where $r_{i,t}(\theta) = \frac{\pi_\theta(y_{i,t}|x,y_{i,<t})}{\pi_{\theta_{\text{old}}}(y_{i,t}|x,y_{i,<t})}$ is the importance ratio. This approach preserves gradient reliability in sparse reward settings while avoiding critic approximation errors.

\paragraph{Decoupled Clip and Dynamic Sampling Policy Optimization (DAPO)} DAPO \citep{DBLP:journals/corr/abs-2503-14476} is an RL approach tailored for reasoning tasks recently. It optimizes the objective as follows:

\begin{small}
\begin{equation}
\begin{aligned}
\mathcal{J}_{\text{DAPO}}(\theta) = \mathbb{E} \left[ \frac{1}{\sum_{i=1}^G |y_i|} \sum_{i=1}^G \sum_{t=1}^{|y_i|} \min \left( r_{i,t}(\theta) \hat{A}_{i,t}, \right. \right. \\
\left. \left. \text{clip}(r_{i,t}(\theta), 1-\epsilon_{\text{low}}, 1+\epsilon_{\text{high}}) \hat{A}_{i,t} \right) \right]
\end{aligned}
\end{equation}
\end{small}

DAPO’s main innovations lie in its decoupled clipping ranges $(1-\epsilon_{\text{low}}, 1+\epsilon_{\text{high}})$, which allow asymmetric policy updates to encourage exploration, dynamic sample filtering that discards batches where all responses share identical correctness, and token-level loss aggregation with reward shaping to handle variations in response lengths.

\subsection{Policy Entropy in Reinforcement Learning}

Policy entropy measures the uncertainty of a policy and reflects the balance between exploration and exploitation in RL \citep{DBLP:conf/icml/AhmedR0S19}. A high-entropy policy encourages diverse outputs and exploration of the action space, whereas a low-entropy policy favors exploiting the currently learned behavior. For a policy model $\pi_\theta$ and a dataset of prompts $\mathcal{D}$, the token-level entropy is:

\begin{small}
\begin{equation}
\mathcal{H}(\pi_\theta, \mathcal{D}) = -\mathbb{E}_{x \sim \mathcal{D}, y \sim \pi_\theta(\cdot|x)} \left[ \frac{1}{|y|} \sum_{t=1}^{|y|} \log \pi_\theta(y_t | y_{<t}, x) \right]
\end{equation}
\end{small}

A major challenge in RL is entropy collapse, in which the policy distribution becomes overly concentrated, resulting in premature convergence, reduced output diversity, and degraded task performance \citep{DBLP:journals/corr/abs-2505-22617}. One common approach to mitigate this issue is to introduce entropy regularization into the policy objective \citep{DBLP:conf/icml/HaarnojaTAL17, DBLP:conf/icml/HaarnojaZAL18}. Specifically, in the context of policy gradient methods, the objective with an entropy regularization term can be written as:

\begin{small}
\begin{equation}
\begin{aligned}
\mathcal{J}(\theta) = \mathbb{E}_{x \sim \mathcal{D}, y \sim \pi_\theta(\cdot|x)} 
&\left[ \sum_{t=1}^{|y|} \hat{A}_t \log \pi_\theta(y_t | y_{<t}, x) \right. \\
&\left. + \alpha \, \mathcal{H}(\pi_\theta(\cdot|x)) \right]
\end{aligned}
\end{equation}
\end{small}

where $\alpha$ is the entropy regularization coefficient that controls the relative strength of the entropy term. A larger $\alpha$ encourages exploration by maintaining higher policy entropy, while a smaller $\alpha$ focuses more on exploitation. In practice, additional strategies such as techniques like the clip-higher method in DAPO \citep{DBLP:journals/corr/abs-2503-14476} can also prevent premature entropy collapse and improve the performance of the policy.

\section{Method}

\subsection{Impact of Clipped-Token Gradients on Entropy Dynamics}

In policy optimization-based RL, PPO and its variants typically employ a clipping operation to constrain the magnitude of policy updates, aiming to stabilize training. Specifically, when a token’s importance sampling ratio exceeds $1+\epsilon$ with a positive advantage, or falls below $1-\epsilon$ with a negative advantage, the corresponding gradient is clipped. While this mechanism effectively prevents overly aggressive updates, it introduces a new issue: policy entropy often becomes unstable during training, typically manifesting as either entropy collapse or entropy explosion. Some existing methods attempt to alleviate this issue by expanding the clipping interval. For example, DAPO’s clip-higher strategy extends the upper bound from $1+\epsilon$ to $1+\epsilon_h$,  incorporating some tokens originally outside the clip interval that contribute higher entropy. This approach primarily addresses entropy collapse, suggesting that gradients from tokens outside the clipping interval play a critical role in controlling the dynamics of policy entropy.

From a theoretical perspective~\citep{Cui2025TheEM}, the change in policy entropy can be approximated as follows, with proof provided in Appendix \ref{appendix-2}.

\begin{small}
\begin{equation}
\begin{aligned}
& \mathcal{H}(\pi_\theta^{k+1}|y_{<t},x) - \mathcal{H}(\pi_\theta^k|y_{<t},x) \\
& \approx -\eta \cdot \text{Cov}_{y \sim \pi_\theta^k(\cdot|x)} \left(\log \pi_\theta^k(y_t|y_{<t},x), \pi_\theta^k(y_t|y_{<t},x) \cdot \hat{A_t} \right)
\end{aligned}
\end{equation}
\end{small}
where $\eta$ denotes the learning rate. This expression shows that the evolution of policy entropy is governed by the covariance between $\log \pi_\theta^k(y_t|y_{<t},x)$ and $\pi_\theta^k(y_t|y_{<t},x) \cdot \hat{A_t}$.

Further analysis indicates that tokens lying outside the clipping interval are predominantly low-probability tokens, as illustrated in Figure \ref{fig:prob_ratio}. Considering their interaction with the advantage function, the effect of these out-of-clip gradients on entropy dynamics can be described more precisely:

\begin{itemize}[leftmargin=*]
    \item For PA\&LP tokens, their gradients would \textbf{encourage the model to explore new possibilities}, reducing the covariance and  slowing the decrease of entropy.
    \item For NA\&LP tokens, their gradients would reinforce high-probability tokens, \textbf{increasing the covariance and accelerating policy convergence}, which leads to a reduction in entropy.
\end{itemize}

These observations demonstrate that gradients from out-of-clip tokens are far from negligible; they directly influence the evolution of policy entropy. 
Appropriately incorporating and regulating out-of-clip gradients enables dynamic control of entropy, guiding the model to achieve a more effective balance between exploration and exploitation at different training stages.

\subsection{Gradient-Preserving Clipping Policy Optimization}

Based on the preceding analysis, we propose that incorporating gradients from tokens outside the clipping interval can effectively control policy entropy. To this end, we introduce the \textbf{Gradient-Preserving Clipping Policy Optimization (CE-GPPO)} algorithm. The core idea of CE-GPPO is to preserve the gradients of tokens outside the clipping interval and adjust their magnitudes while ensuring stable policy updates, enabling explicit regulation of entropy. Specifically, CE-GPPO decouples the forward and backward passes by introducing a \emph{stop gradient} operation, allowing gradient updates to no longer be strictly constrained by the original clipping interval. The objective function is defined as:

\begin{small}
\begin{equation}
\begin{aligned}
\label{grpo_loss_soft_general}
&\mathcal{J}_{\text{CE-GPPO}}(\theta) 
= \mathbb{E} 
   \left[\frac{1}{\sum_{i=1}^G |y_i|} 
   \sum_{i=1}^G \sum_{t=1}^{|y_i|} 
   \ell^{(i)}\right] ,
\\[3pt]
&\ell^{(i)} =
\begin{cases}
\beta_1 \cdot \, \dfrac{1-\epsilon}{\operatorname{sg}(\delta)} 
         \, \delta \, \cdot \hat{A}_{i,t}, 
& \text{if } \delta < 1-\epsilon \text{ and } \hat{A}_{i,t} < 0, \\[8pt]
\beta_2 \cdot \, \dfrac{1+\epsilon}{\operatorname{sg}(\delta)} 
         \, \delta \, \cdot \hat{A}_{i,t}, 
& \text{if } \delta > 1+\epsilon \text{ and } \hat{A}_{i,t} > 0 , \\[8pt] 
\delta \cdot \, \hat{A}_{i,t}, 
& \text{otherwise}.
\end{cases}
\end{aligned}
\end{equation}
\end{small}
Here, $\delta=r_{i,t}$ denotes the importance sampling ratio, and $\operatorname{sg}(\cdot)$ represents the \emph{stop gradient} operation. The coefficients $\beta_1$ and $\beta_2$ control the scaling of gradients outside the left and right clipping boundaries, respectively. It is worth noting that when $\beta_1=\beta_2=0$, CE-GPPO is equivalent to PPO.

Through this design, CE-GPPO effectively incorporates the gradients of originally clipped PA\&LP and NA\&LP tokens into the update. Specifically, a larger $\beta_1$ amplifies PA\&LP gradients, slowing the decline of policy entropy and promoting exploration, whereas a larger $\beta_2$ amplifies NA\&LP gradients, accelerating policy convergence and reducing entropy to facilitate exploitation. By adjusting these coefficients, CE-GPPO can flexibly modulate the dynamics of policy entropy,  balancing exploration and exploitation during training.

\subsection{Ensuring Stable Optimization in CE-GPPO}

Despite incorporating gradients from outside the clipping interval to regulate policy entropy, CE-GPPO maintains a stable optimization process. This property can be understood by analyzing its gradient formulation. The gradient of CE-GPPO is given as follows, with a detailed derivation provided in Appendix~\ref{appendix:ce-gppo-gradient}.

\begin{small}
\begin{equation}
\begin{aligned}
\nabla_\theta \mathcal{J}_{\text{CE-GPPO}}(\theta) &= 
\mathbb{E} \left[
\frac{1}{\sum_{i=1}^G |y_i|} \sum_{i=1}^G \sum_{t=1}^{|y_i|}
\mathcal{F}_{i,t}(\theta) \cdot \right. \\
&\left. \nabla_\theta \log \pi_\theta(y_{i,t}| y_{<t}, x) \cdot \hat{A}_{i,t}
\right]
\end{aligned}
\end{equation}
\end{small}


\begin{small}
\begin{equation}
\begin{aligned}
\mathcal{F}_{i,t}(\theta) &= 
\begin{cases}
\beta_1 \cdot (1-\epsilon), & \text{if } \delta < 1-\epsilon \text{ and } \hat{A}_{i,t} < 0, \\[0.5em]
\beta_2 \cdot (1+\epsilon), & \text{if } \delta > 1+\epsilon \text{ and } \hat{A}_{i,t} > 0, \\[0.5em]
\delta, & \text{otherwise}.
\end{cases}
\end{aligned}
\end{equation}
\end{small}

It can be observed that when the importance sampling ratio 
$\delta$ falls outside the clipping interval, CE-GPPO does not amplify the gradient without bound. Instead, it restricts the update to $\beta_1 \cdot (1-\epsilon)$ or $\beta_2 \cdot (1+\epsilon)$. Since $\beta_1$ and $\beta_2$ are typically close to 1, the overall gradient magnitude remains within a reasonable range. Moreover, other terms are structurally identical to those in the standard PPO gradient, and therefore contribute to optimization stability in the same way as the original PPO. As a result, while CE-GPPO introduces gradient signals beyond the clipping interval, it still preserves stability comparable to that of standard PPO,  ensuring a controllable optimization process.

In addition, CE-GPPO preserves the same computational and memory complexity as standard PPO, since it only introduces an element-wise reweighting of the policy gradient loss without additional forward passes or auxiliary components.

\section{Experiment}

\subsection{Experimental Setup}

\paragraph{Datasets}


Our RL training dataset is KlearReasoner-MathSub-30K~\citep{DBLP:journals/corr/abs-2508-07629}, which consists of approximately 30k samples collected from several high-quality sources, including Skywork-OR1 \citep{DBLP:journals/corr/abs-2505-22312}, Acereason \citep{DBLP:journals/corr/abs-2505-16400}, NuminaMath \citep{numina_math_datasets}, and DeepScaleR \citep{deepscaler2025}. To mitigate potential data contamination, the dataset has been further processed with 9-gram deduplication against the evaluation benchmarks.

\paragraph{Training}


We conducted training with CE-GPPO on two model sizes, DeepSeek-R1-Distill-Qwen-1.5B and DeepSeek-R1-Distill-Qwen-7B. The maximum training sequence length was set to 16k, and the learning rate was fixed at $1\times 10^{-6}$. We generated 8 rollouts for each prompt. The parameter $\epsilon$ was fixed at 0.2 with symmetric upper and lower bounds. Following \citet{DBLP:journals/corr/abs-2505-22312}, no KL loss term was included in the objective. Each run was trained for up to 1000 steps, corresponding to approximately 10 epochs. The experimental settings for the baselines are provided in Appendix \ref{appendix-1}.

\paragraph{Evaluation}



We evaluated our method on multiple open-ended benchmarks covering mathematical reasoning, code reasoning, and instruction following. For the mathematics benchmarks, all results were reported using avg@32, except for MATH500, which used avg@4. For AIME 24/25, we conducted inference with a maximum sequence length of 32k, while all other mathematics benchmarks were evaluated with a maximum sequence length of 16k. Following \citet{yang2024qwen2}, answers were extracted from the \verb|\boxed{}| in the model's output. For the code reasoning and instruction-following benchmarks, all results were reported using avg@4, with a maximum inference length of 32k. In particular, for the instruction-following benchmarks, we reported results at both the prompt level (IFBench-P) and the instance level (IFBench-I).


\begin{table*}[t]
\centering
\resizebox{1.0\linewidth}{!}{ 
\begin{tabular}{lccccccccc}
\toprule
\multirow{2}{*}{\textbf{Method}} 
& \multicolumn{5}{c}{\textbf{Math Reasoning}} 
& \multicolumn{2}{c}{\textbf{Code Reasoning}} 
& \multicolumn{2}{c}{\textbf{Instruct Following}} \\
\cmidrule(lr){2-6}\cmidrule(lr){7-8}\cmidrule(lr){9-10}
& \textbf{AIME24} & \textbf{AIME25} & \textbf{HMMT25} & \textbf{MATH500} & \textbf{AMC23} 
& \textbf{HumanEval} & \textbf{LCB v6} 
& \textbf{IFBench-P} & \textbf{IFBench-I} \\
\midrule
\textbf{DS-R1-Distill-Qwen-1.5B} & 29.2 & 24.1 & 13.1 & 86.0 & 73.7 & 70.4 & 25.1 & 12.0 & 14.1 \\
\quad + GRPO  & 33.4 & 28.1 & 16.6 & 88.3 & 79.3 & 67.5 & 27.1 & 12.2 & 14.5 \\
\quad + DAPO  & 40.0 & 28.4 & 19.2 & 90.0 & 84.4 & 73.2 & 30.5 & 12.8 & 14.8 \\
\rowcolor{gray!15} 
\quad + CE-GPPO w/ $\beta_1=0.5,\beta_2=1$ & 42.0 & \textbf{33.9} & \textbf{21.6} & \textbf{91.0} & \textbf{85.9} & \textbf{76.5} & \textbf{31.7} & 13.7 & 15.7 \\
\rowcolor{gray!15} 
\quad + CE-GPPO w/ $\beta_1=0.75,\beta_2=1$ & \textbf{43.6} & 31.0 & 19.3 & 90.9 & 85.6 & 74.9 & 31.1 & \textbf{13.8} & \textbf{16.0} \\
\midrule
\textbf{DS-R1-Distill-Qwen-7B} & 54.5 & 39.1 & 26.2 & 93.6 & 90.6 & 89.6 & 49.0 & 16.8 & 18.9 \\
\quad + GRPO  & 55.3 & 40.3 & 24.5 & 93.7 & 88.8 & 88.6 & 49.2 & 16.6 & 18.9 \\
\quad + DAPO  & 59.7 & 48.7 & 25.6 & 95.1 & 93.4 & 92.5 & 52.2 & 16.5 & 18.7 \\
\rowcolor{gray!15} 
\quad + CE-GPPO w/ $\beta_1=0.5,\beta_2=1$ & 62.3 & 49.1 & 27.3 & 94.9 & 92.8 & 91.9 & 52.2 & \textbf{17.4} & \textbf{19.8} \\
\rowcolor{gray!15} 
\quad + CE-GPPO w/ $\beta_1=0.75,\beta_2=1$ & \textbf{66.0} & \textbf{51.4} & \textbf{30.5} & \textbf{95.6} & \textbf{93.8} & \textbf{93.0} & \textbf{53.6} & \textbf{17.4} & 19.7 \\
\bottomrule
\end{tabular}
}
\caption{Performance comparison of CE-GPPO and baseline methods on multiple benchmarks across different models. DS-R1-Distill-Qwen-1.5B and DS-R1-Distill-Qwen-7B denote the DeepSeek-R1-Distill-Qwen-1.5B and DeepSeek-R1-Distill-Qwen-7B models, respectively. DAPO stands for GRPO training with Clip Higher trick.}
\end{table*}

\begin{figure}[t]
    \centering
    \begin{subfigure}{0.48\linewidth}
        \centering
        \includegraphics[width=\linewidth]{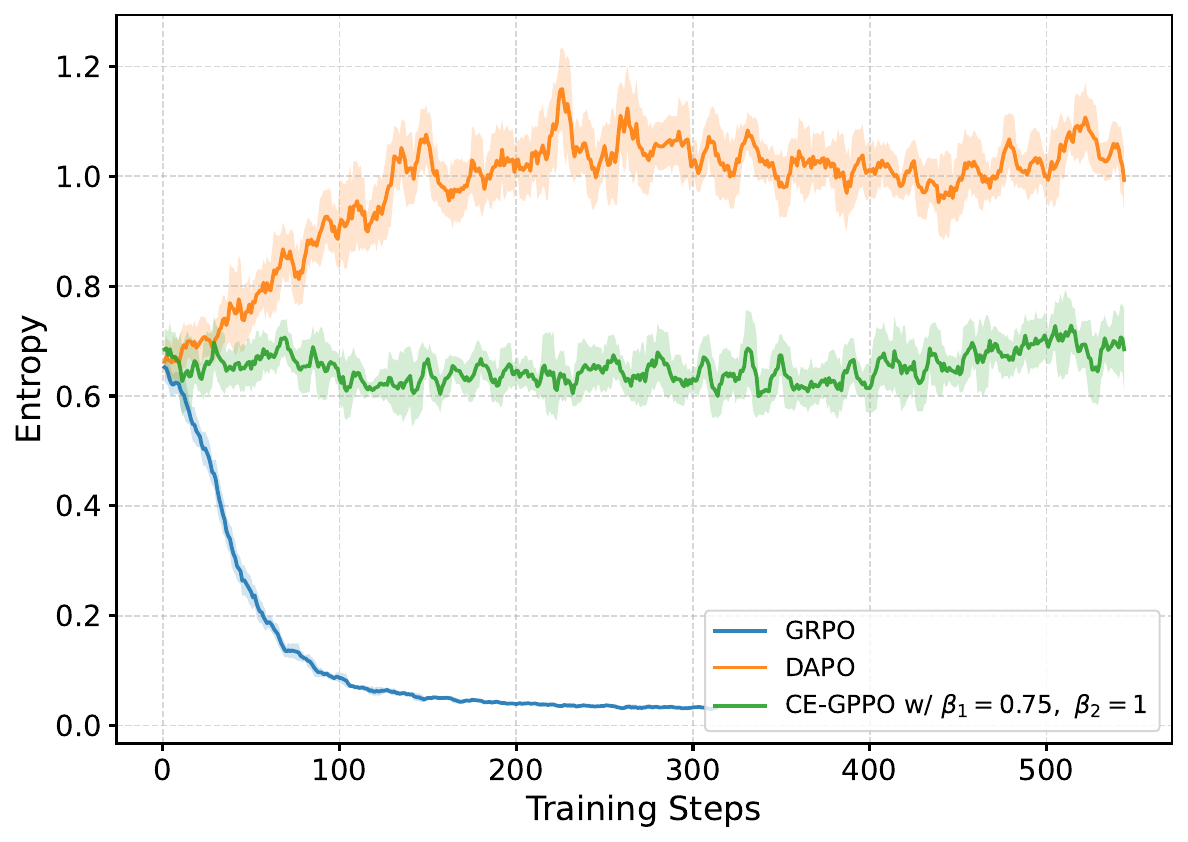}
        \caption{Entropy Dynamics}
        \label{fig:main_entropy_dymaic}
    \end{subfigure}
    \hfill
    \begin{subfigure}{0.48\linewidth}
        \centering
        \includegraphics[width=\linewidth]{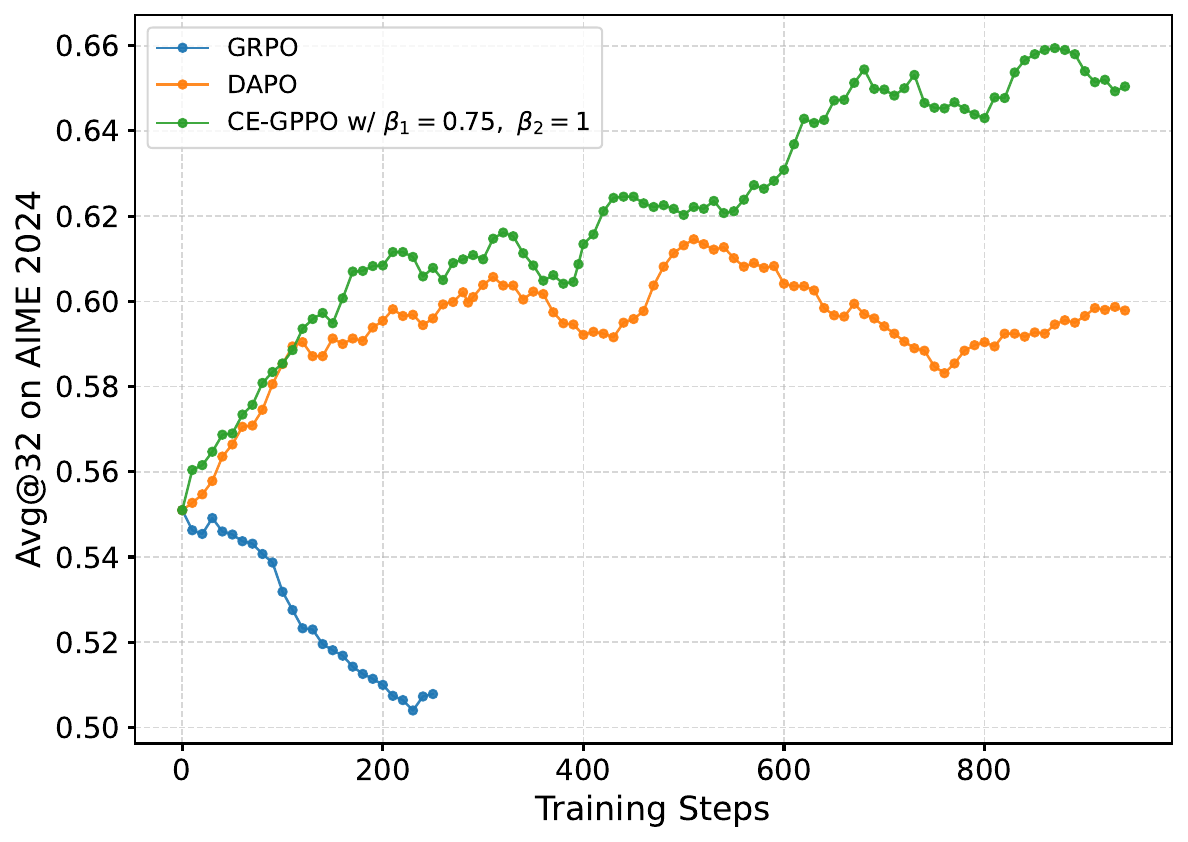}
        \caption{Benchmark Accuracy}
        \label{fig:main_acc_dymaic}
    \end{subfigure}
    \caption{Based on DeepSeek-R1-Distill-Qwen-7B, a comparison of GRPO, DAPO, and GPPO in terms of entropy dynamics and AIME25 benchmark accuracy.}
    \label{fig:main_dymaic}
\end{figure}

\subsection{Main Results}


As shown in Table 1, we report the performance comparison between our proposed CE-GPPO and other baseline methods. DAPO represents GRPO training with the clip-higher strategy, where we set $\epsilon_{\text{high}}$ to $0.28$ with reference to ~\citet{DBLP:journals/corr/abs-2503-14476}.  It can be observed that CE-GPPO consistently outperforms the baselines across different benchmarks, with particularly pronounced gains on more challenging tasks such as AIME25 and HMMT25. Moreover, the advantages of CE-GPPO scale with model size: the 1.5B model achieves a 2.5-point improvement over the best baseline, while the 7B model achieves a 3-point improvement. This indicates that larger models can more effectively leverage the benefits of CE-GPPO.

Furthermore, as shown in Figure~\ref{fig:main_dymaic}, we examine the training dynamics of entropy and AIME24 accuracy, and obtain three key observations.  
\begin{itemize}[leftmargin=*]
    \item First, native GRPO suffers from entropy collapse, an issue effectively mitigated by both DAPO and CE-GPPO, leading to significant improvements. DAPO addresses this by adjusting Clip Higher, whereas CE-GPPO propagates the gradients of clipped PA\&LP tokens back in a bounded and moderate manner.
    \item Second, at the early stage of DAPO training, entropy increases significantly and remains at a high level, while CE-GPPO maintains stable entropy throughout training and achieves clear improvements over DAPO. This suggests that DAPO is prone to over-exploration, whereas CE-GPPO resolves this problem by backpropagating the gradients of clipped NA\&LP tokens.
    \item Third, by coordinating entropy dynamics, CE-GPPO achieves a stable balance between exploration and exploitation, leading to more effective optimization.
\end{itemize}



\section{Analysis}

\subsection{Impact of Different $\beta$ Hyperparameters on Entropy Dynamics}

\begin{figure}[t]
    \centering
    \begin{subfigure}{0.48\linewidth}
        \centering
        \includegraphics[width=\linewidth]{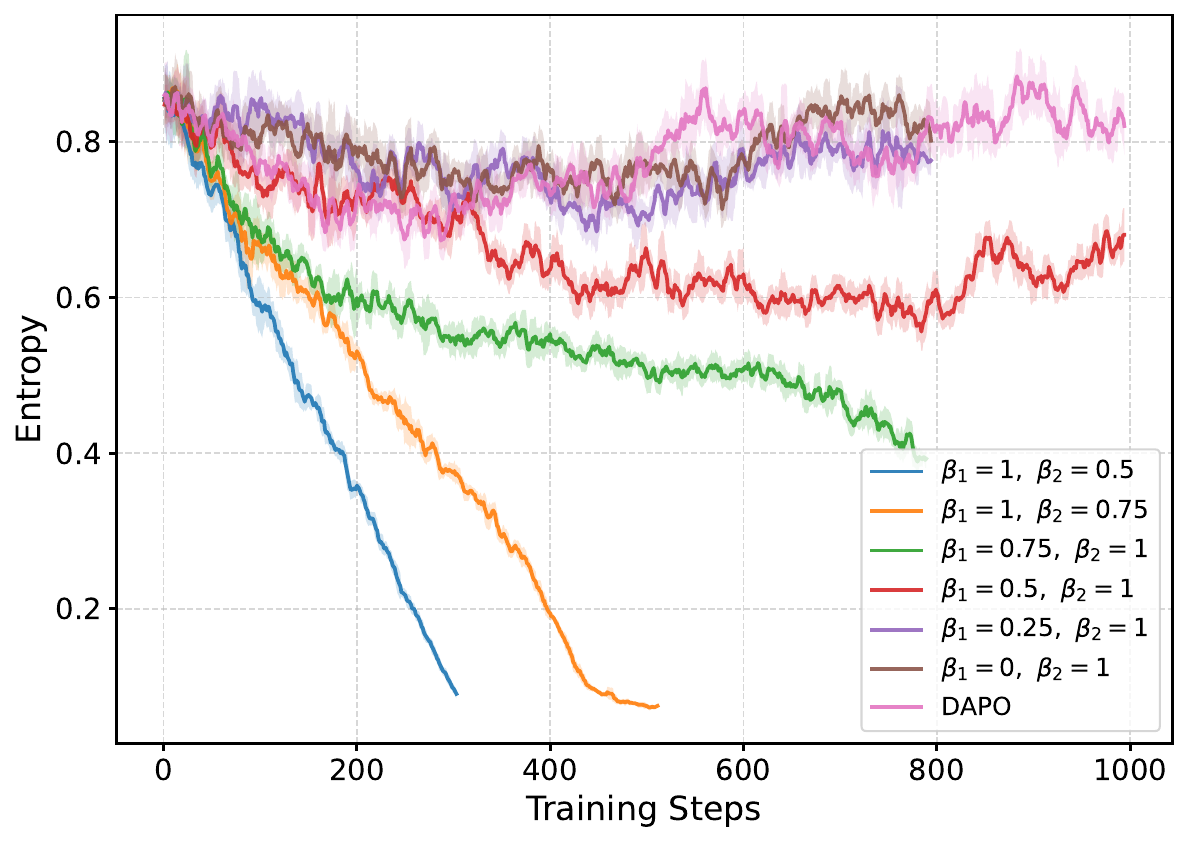}
        \caption{Entropy Dynamics (1.5B)}
        \label{fig:entropy_1}
    \end{subfigure}
    \hfill
    \begin{subfigure}{0.48\linewidth}
        \centering
        \includegraphics[width=\linewidth]{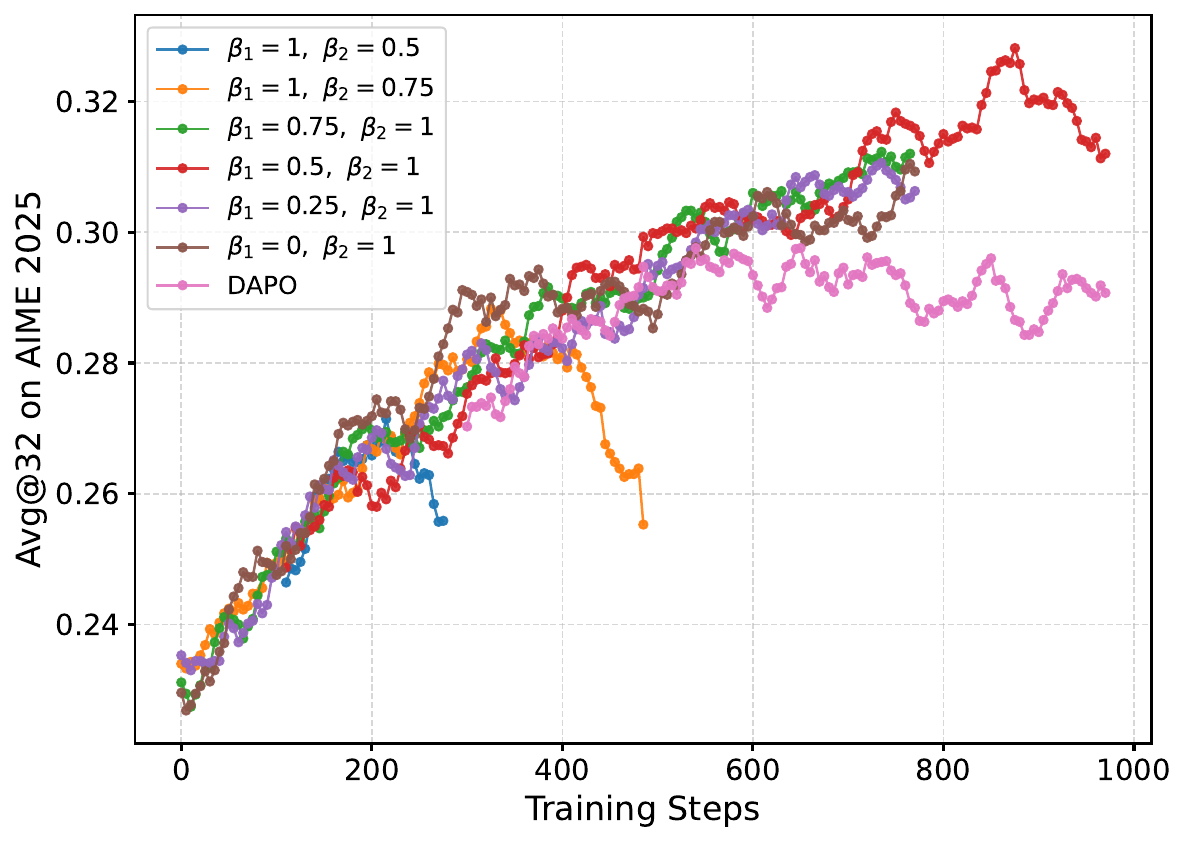}
        \caption{Benchmark Acc (1.5B)}
        \label{fig:entropy_2}
    \end{subfigure}

    \vspace{0.3cm} 

    \begin{subfigure}{0.48\linewidth}
        \centering
        \includegraphics[width=\linewidth]{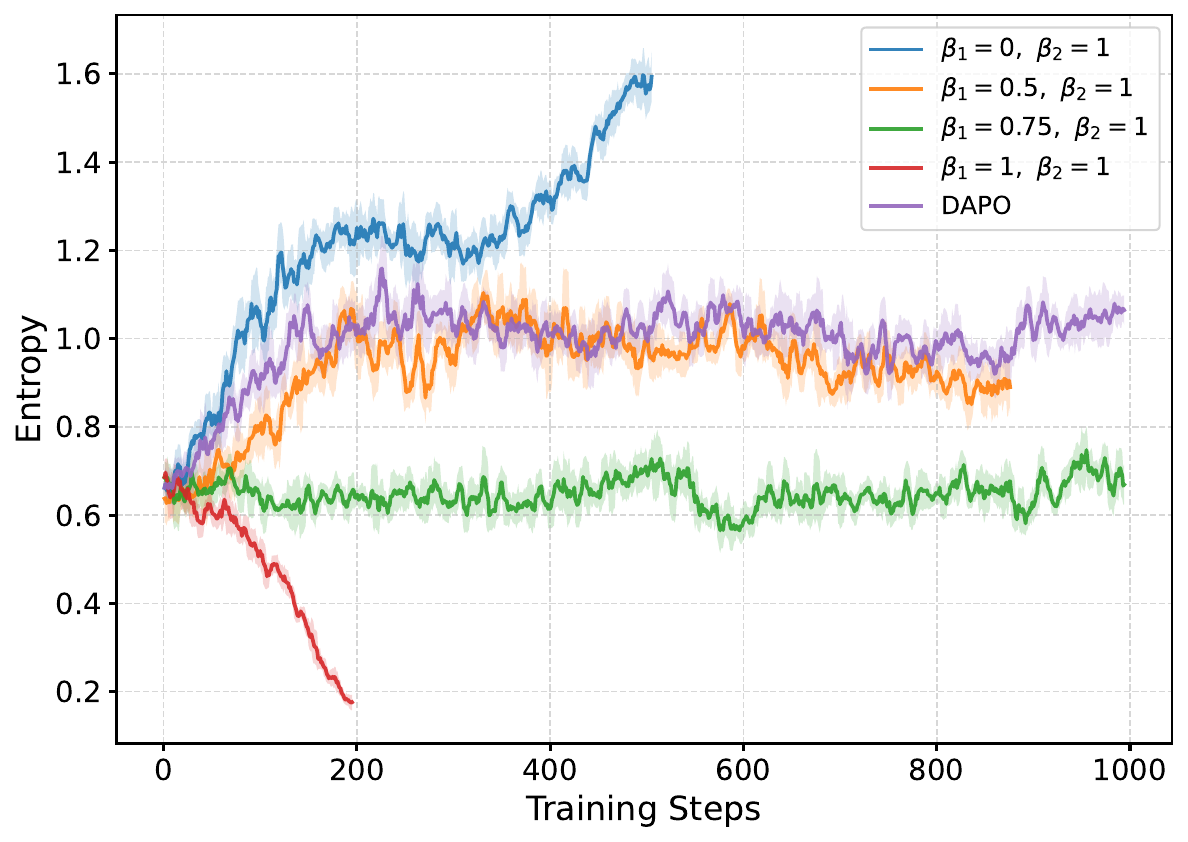}
        \caption{Entropy Dynamics (7B)}
        \label{fig:entropy_3}
    \end{subfigure}
    \hfill
    \begin{subfigure}{0.48\linewidth}
        \centering
        \includegraphics[width=\linewidth]{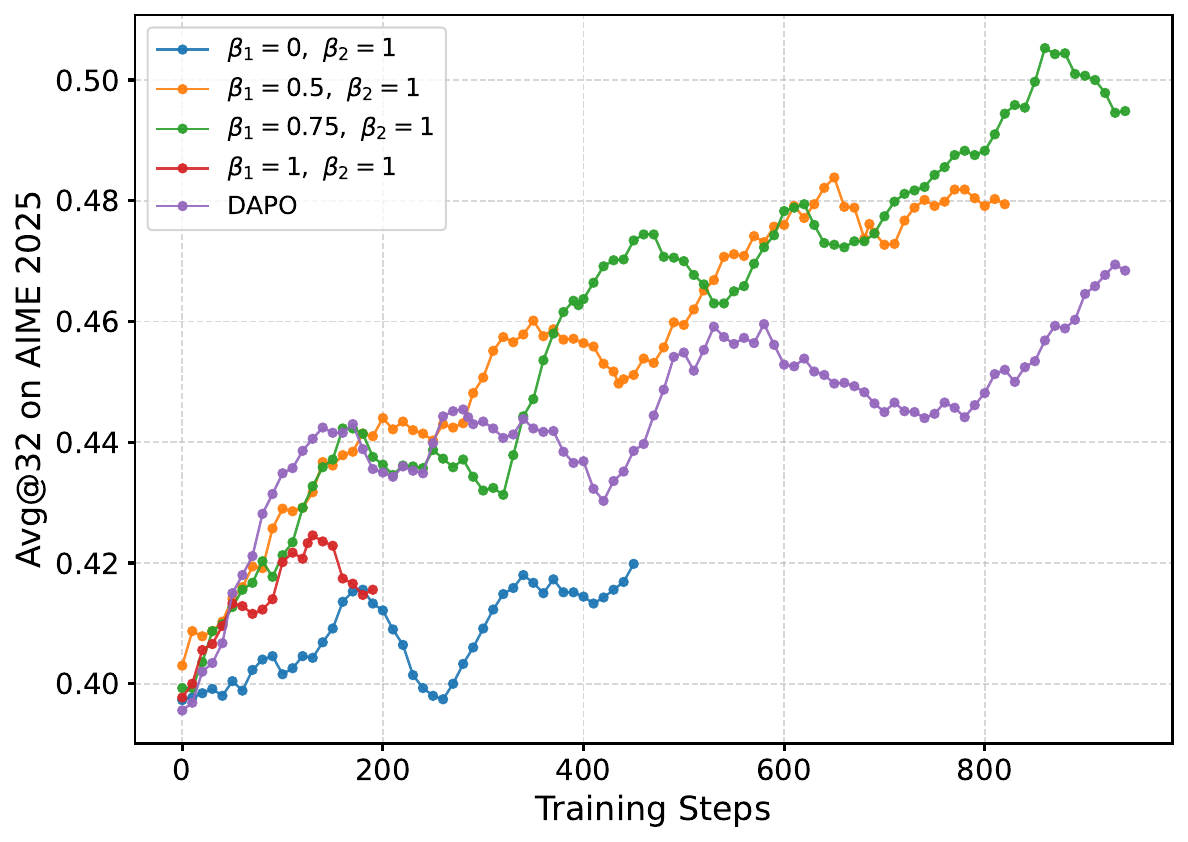}
        \caption{Benchmark Acc (7B)}
        \label{fig:entropy_4}
    \end{subfigure}

    \caption{Entropy dynamics and benchmark accuracy under different $\beta_1/\beta_2$ configurations.}
    \label{fig:entropy_1_4}
\label{fig:fig2}
\end{figure}





To empirically validate that CE-GPPO can regulate entropy dynamics through parameter settings, we conducted experiments with different $\beta_1/\beta_2$ configurations, as shown in Figure \ref{fig:fig2}. We observe that entropy decreases much faster when $\beta_1$ is large or $\beta_2$ is small, whereas it decreases much more slowly when $\beta_1$ is small or $\beta_2$ is large. This confirms our key finding that the choice of $\beta_1$ and $\beta_2$ directly governs the evolution of entropy. The underlying mechanism is that:

\begin{itemize}[leftmargin=*]
    \item A larger $\beta_1$ amplifies gradients beyond the left clip boundary (mainly from NA\&LP tokens). These gradients strengthen high-probability tokens, accelerating exploitation and thus causing entropy to collapse quickly.
    \item A larger $\beta_2$, on the other hand, amplifies gradients beyond the right clip boundary (mainly from PA\&LP tokens). These gradients encourage exploration of new tokens,  slowing entropy reduction.
\end{itemize}

\subsection{Entropy-Guided Training Dynamics}





Further experiments reveal the relationship between entropy dynamics and model performance. As shown in Figure ~\ref{fig:fig2}, when entropy decreases too rapidly, such as in the 1.5B model with the setting $\beta_1=1,\beta_2=0.5$, the model performance degrades quickly once entropy falls below a certain threshold. In contrast, when entropy remains relatively high and stable in the early stage of training, as in the 1.5B model with the setting $\beta_1=0.5,\beta_2=1$, the model avoids premature convergence to suboptimal solutions and continues to improve on benchmarks.

However, excessively high entropy does not necessarily lead to better results, and it must be maintained within a reasonable range. In the 7B model with the setting $\beta_1=0,\beta_2=1$, entropy increases consistently during the early stage of training, yet model performance does not show clear gains. Instead, the configuration $\beta_1=0.75,\beta_2=1$, which keeps entropy more stable, achieves the best performance throughout training.
Although the entropy dynamics of the 1.5B and 7B models differ, both have reached three consistent conclusions:

\begin{itemize}[leftmargin=*]
    \item Maintaining \textbf{relatively high and stable entropy} is generally beneficial for sustained performance improvement during training.
    \item \textbf{A greater weight $\beta_2$ is given to the gradients of PA\&LP tokens, while a smaller weight $\beta_1$ is given to the gradients of NA\&LP tokens}. This is conducive to maintaining the model's exploration ability and is more beneficial for performance improvement.
    \item We have observed that CE-GPPO exhibits \textbf{robustness to hyperparameters}, with significant performance improvements seen across models of different sizes under the settings of \(\beta_1=0.5, \beta_2=1\) and \(\beta_1=0.75, \beta_2=1\).
\end{itemize}
Moreover, we further investigate the role of entropy at different training stages in section ~\ref{the_role_entropy_stage}. Similar to the findings of ~\citet{DBLP:journals/corr/abs-2507-20534}, we find that maintaining high entropy in the early stage and moderately reducing entropy in the later stage facilitates better performance improvement.

\subsection{Training Stability Analysis of CE-GPPO}

\begin{figure}[t]
    \centering
    \begin{subfigure}{0.48\linewidth}
        \centering
        \includegraphics[width=\linewidth]{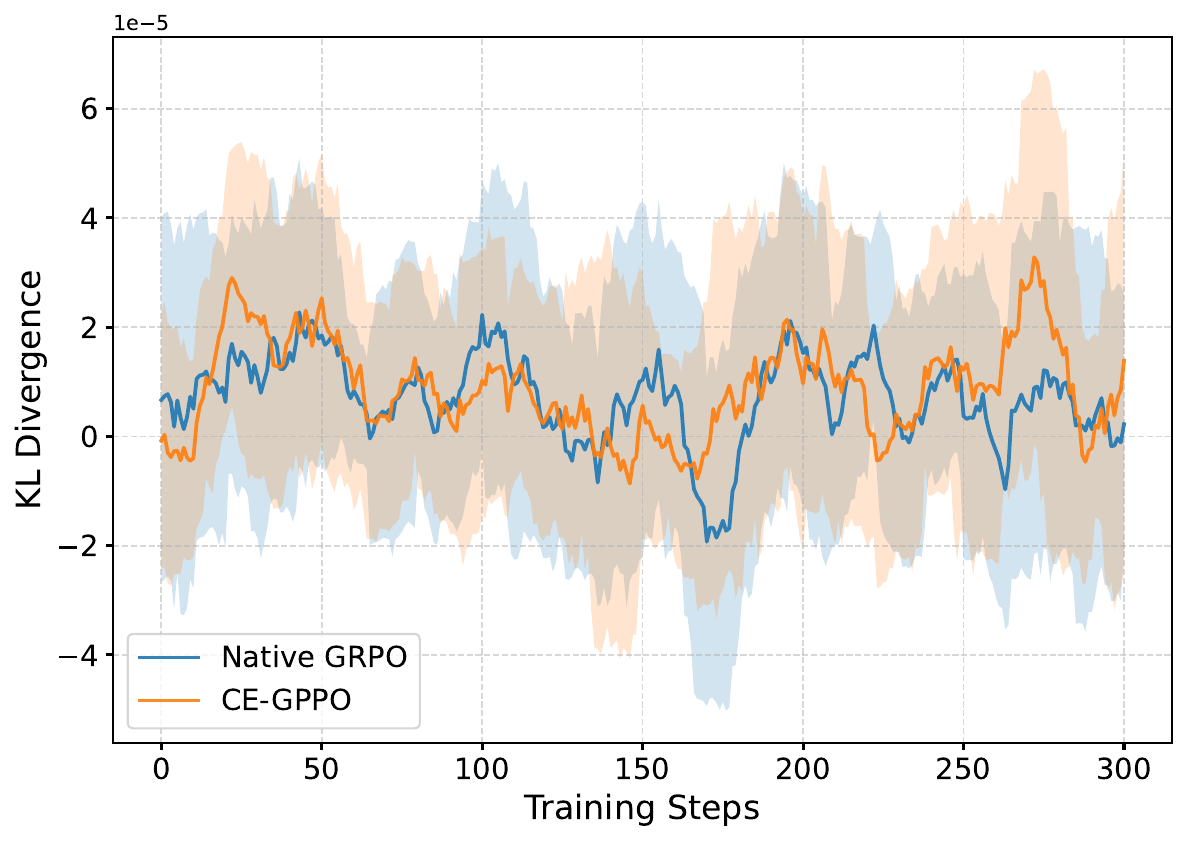}
        \caption{KL Divergence}
        \label{fig:kl}
    \end{subfigure}
    \hfill
    \begin{subfigure}{0.48\linewidth}
        \centering
        \includegraphics[width=\linewidth]{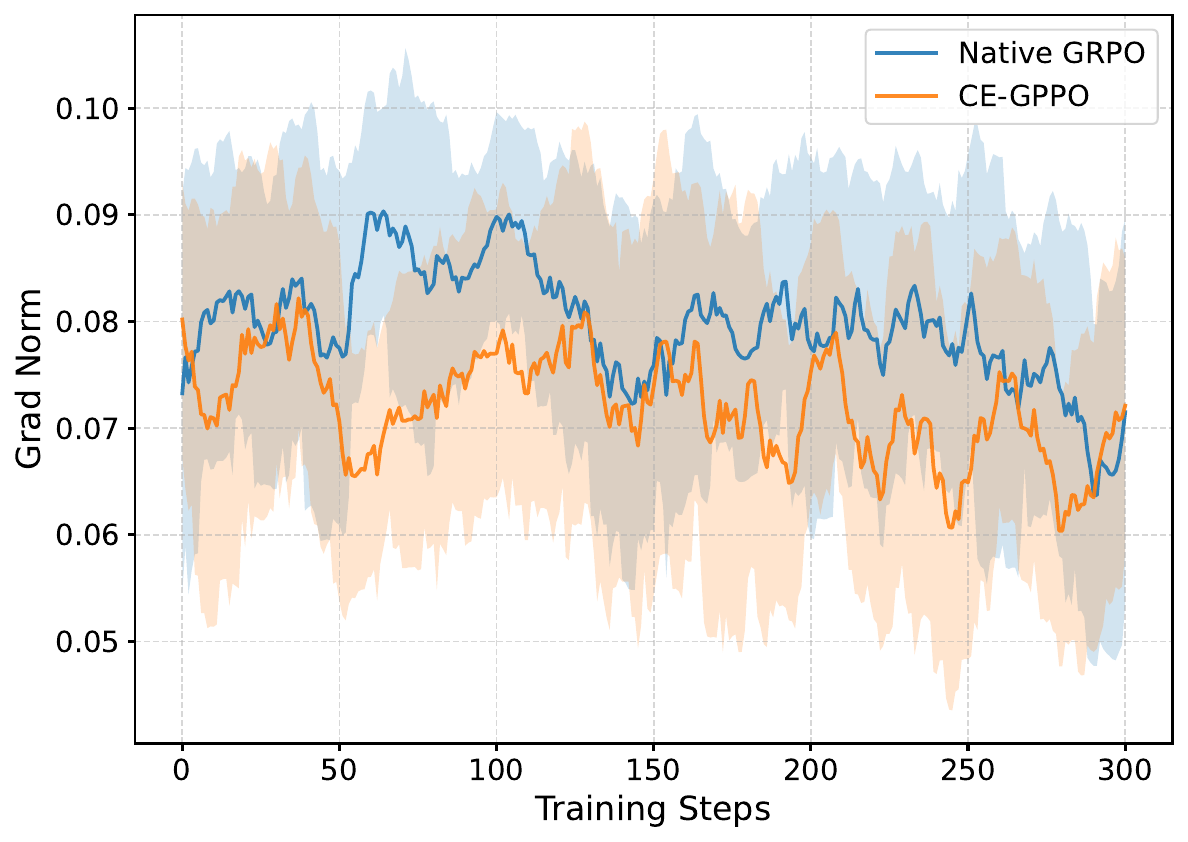}
        \caption{Gradient Norm}
        \label{fig:grad_norm}
    \end{subfigure}
    \caption{Comparison of KL divergence and gradient norm dynamics between GRPO and CE-GPPO.}
    \label{fig:kl_and_grad_norm}
\end{figure}

To achieve finer control over policy entropy, CE-GPPO introduces the gradients of tokens outside the clipping interval on top of GRPO. Although this operation to some extent relaxes the trust region constraint of standard PPO, we theoretically prove that the additional gradients incorporated by CE-GPPO are stable and do not lead to training collapse. To further validate this conclusion, we compare GRPO and CE-GPPO in terms of the KL divergence between the old policy model and the policy model during training, as well as the variation in gradient norms, as shown in Figure \ref{fig:kl_and_grad_norm}. The results show that CE-GPPO maintains a stable trend in both metrics throughout training, without abrupt fluctuations or abnormal values beyond reasonable ranges. These findings provide empirical evidence that CE-GPPO backpropagates gradients of out-of-trust-region tokens in a mild and bounded manner, preventing the policy model from drifting too far from the old policy model, and ensuring that CE-GPPO training remains stable.

\subsection{Comparison with Other RL Algorithms}

\begin{table*}[t]
\centering
\resizebox{0.9\linewidth}{!}{ 
\begin{tabular}{lcccccc}
\toprule
\textbf{Method} & \textbf{AIME24} & \textbf{AIME25} & \textbf{HMMT25} & \textbf{MATH500} & \textbf{AMC23} & \textbf{Avg.} \\
\midrule
\textbf{DS-R1-Distill-Qwen-1.5B} & 29.2 & 24.1 & 13.1 & 86.0 & 73.7 & 45.2 \\
\quad + CISPO  & 32.9 & 25.1 & 13.2 & 85.8 & 80.9 & 47.6 \\
\quad + GSPO  & \textbf{42.5} & 33.6 & 19.0 & 90.3 & \textbf{85.9} & 54.3 \\
\rowcolor{gray!15} 
\quad + CE-GPPO w/ $\beta_1=0.5,\beta_2=1$ & 42.0 & \textbf{33.9} & \textbf{21.6} & \textbf{91.0} & \textbf{85.9} & \textbf{54.9} \\
\bottomrule
\end{tabular}
}
\caption{ Comparison of CE-GPPO, CISPO and GSPO in mathematical RL training.}
\label{Other_RL}
\end{table*}

In this section, we compare CE-GPPO with a broader set of RL algorithms, including CISPO~\citep{DBLP:journals/corr/abs-2507-18071}, and GSPO~\citep{DBLP:journals/corr/abs-2507-18071}. The experimental configurations  are provided in Appendix \ref{appendix-1}. As summarized in Table ~\ref{Other_RL}, we evaluate each model on several mathematical reasoning benchmarks. CE-GPPO achieves the best performance on $4$ out of $5$ datasets, demonstrating significant improvements over the baseline methods and underscoring the effectiveness of the proposed approach.

We observe that CISPO exhibits model collapse during training: in later stages, its performance declines sharply alongside a rapid drop in entropy. CISPO also retains gradients from all tokens while applying constraints on the gradient magnitudes. This suggests that constraining gradient norms alone is insufficient to ensure training stability when gradients from all tokens are retained.
In contrast, CE-GPPO shows steady improvement throughout training. We identify two main reasons for this robustness:
\begin{itemize}[leftmargin=*]
\item CE-GPPO assigns a smaller weight $\beta_1$ to the gradients of NA\&LP tokens, while assigning a larger weight $\beta_2$ to the gradients of PA\&LP tokens. Compared with CISPO, it achieves a better balance between exploration and exploitation.
\item Compared with CISPO, CE-GPPO inherits the pessimistic update mechanism of PPO. Specifically, when $\delta < 1-\epsilon_l$ and $\hat{A}_{i,t} > 0$, CE-GPPO sets $\mathcal{F}_{j,t}(\theta) = \delta$ (where $0 < \delta < 1-\epsilon_l$), suppressing overly optimistic improvements and thus updating more conservatively; whereas CISPO uses a larger update magnitude $1-\epsilon_l$. When $\delta > 1+\epsilon_l$ and $\hat{A}_{i,t} < 0$, CE-GPPO still sets $\mathcal{F}_{j,t}(\theta) = \delta$ (with $\delta > \epsilon_h$), fully trusting negative feedback without suppression, while CISPO applies a smaller update magnitude $1+\epsilon_l$. The pessimistic update strategy of CE-GPPO avoids excessive optimism while fully incorporating negative feedback, enhancing algorithmic stability and preventing policy collapse.
\end{itemize}

Compared with GSPO, CE-GPPO shows clear advantages on AIME2025, HMMT25 and MATH500. It also achieves a higher average score. We attribute this improvement to CE-GPPO’s ability to preserve gradients across more tokens. During GSPO training, nearly 15\% of tokens are clipped and do not contribute to gradient updates. Thus, CE-GPPO not only achieves better performance but also higher token utilization efficiency.

Importantly, CE-GPPO and GSPO introduced complementary improvements. GSPO replaced token-level importance sampling ratios with sequence-level ratios to reduce variance, while CE-GPPO modified the policy gradient objective via stop-gradient weighting, enabling controlled gradient contributions beyond the clipping region. Prior work has shown that GSPO may suffer from entropy explosion \cite{entropy_preserving_rl_iclr2026}, suggesting that variance reduction alone does not resolve instability in entropy dynamics, an issue that CE-GPPO explicitly addresses.







\begin{figure}[t]
    \centering
    \begin{subfigure}{0.48\linewidth}
        \centering
        \includegraphics[width=\linewidth]{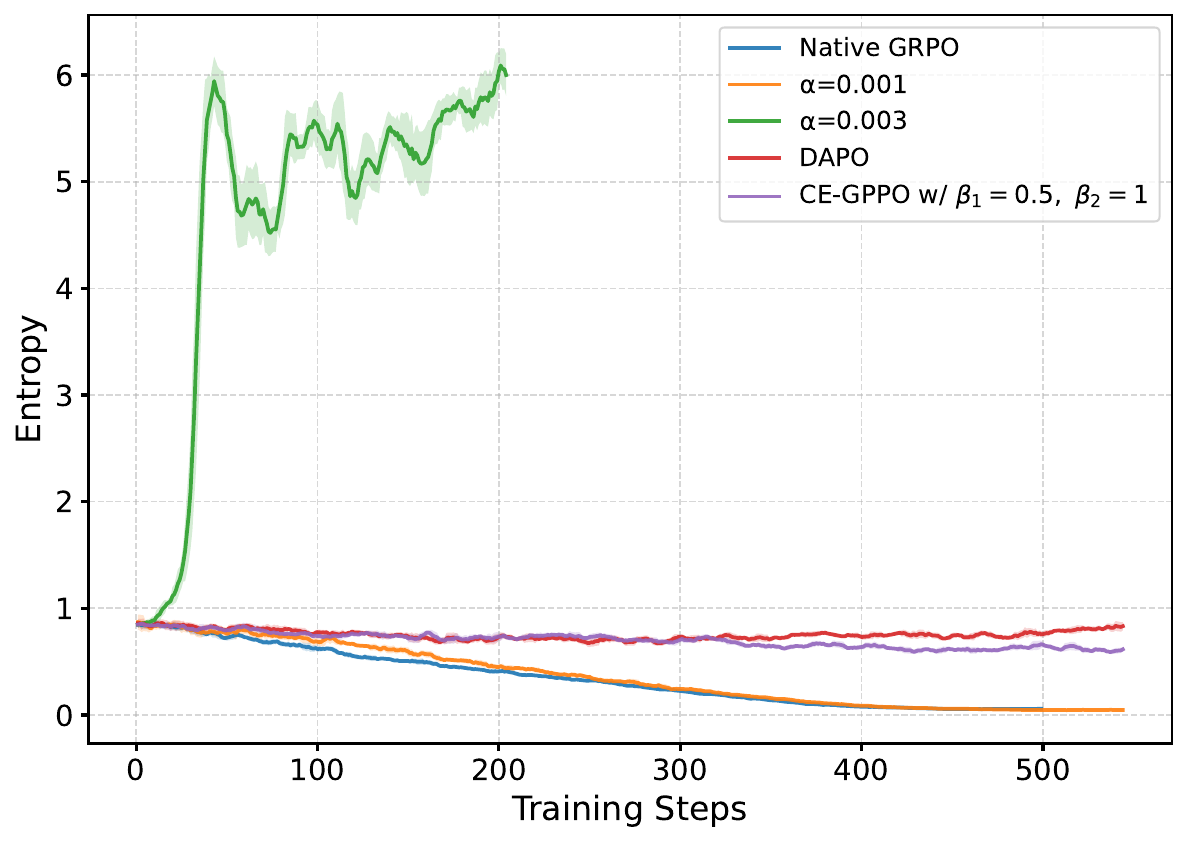}
        \caption{Entropy Dynamics}
        \label{fig:entropy_loss}
    \end{subfigure}
    \hfill
    \begin{subfigure}{0.48\linewidth}
        \centering
        \includegraphics[width=\linewidth]{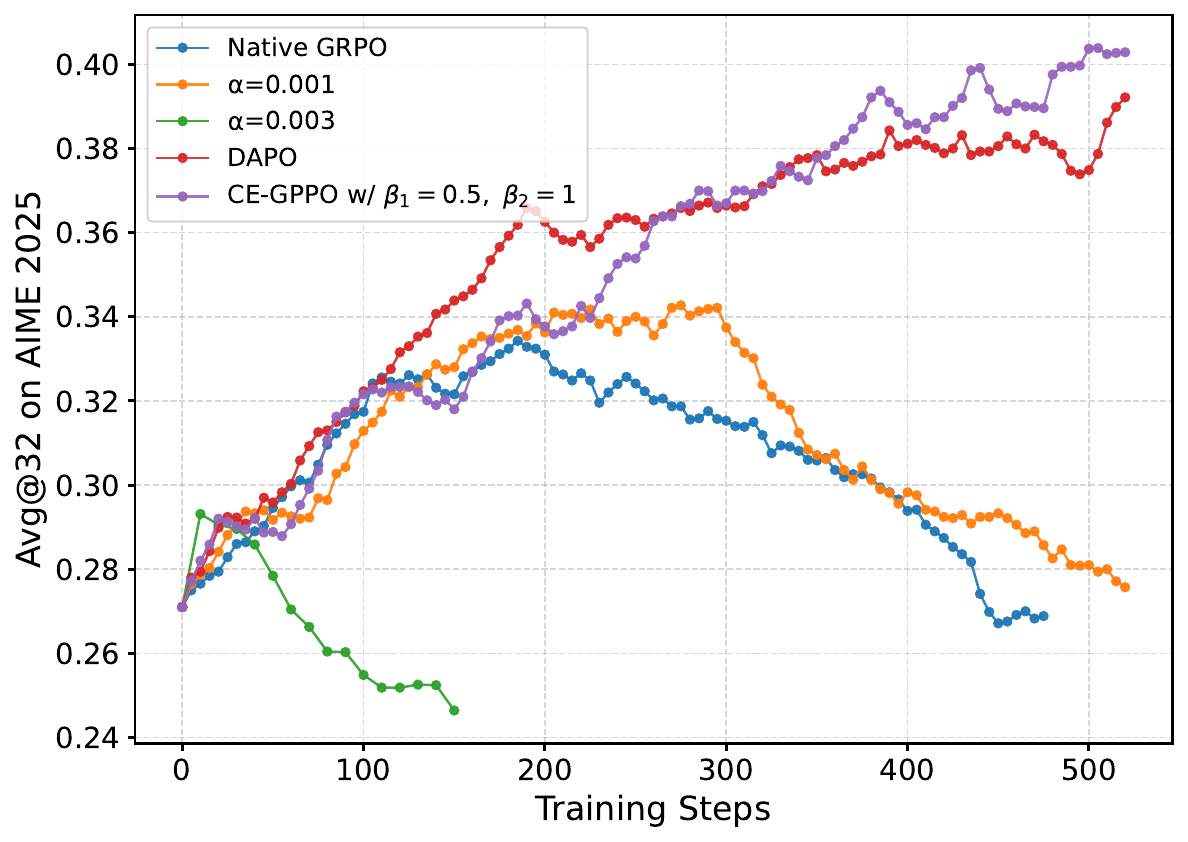}
        \caption{Benchmark Accuracy}
        \label{fig:grad_norm}
    \end{subfigure}
    
    \caption{Comparison of CE-GPPO with other entropy collapse mitigation strategies. Native GRPO denotes the baseline without any mitigation strategy. $\alpha = 0.001/0.003$ indicate the addition of an entropy loss term to the Native GRPO baseline, where $\alpha$ represents the entropy loss coefficient. DAPO refers to applying the Clip Higher strategy on Native GRPO baseline.}
    \label{fig:entropy_loss}
\end{figure}

\subsection{Comparison with Existing Entropy Collapse Mitigation Methods}

In this section, we compare CE-GPPO with other methods designed to mitigate entropy collapse, focusing on the relationship between entropy dynamics and model performance. As shown in Figure ~\ref{fig:entropy_loss}, the Native GRPO without any entropy collapse mitigation strategy suffers from severe entropy collapse during training, with entropy eventually converging to around 0.06. Both the clip higher strategy adopted by DAPO and the traditional entropy regularization method can slow down entropy decay in RL to some extent.

However, we find that entropy regularization is highly sensitive to the choice of its coefficient. Setting the entropy loss coefficient to $\alpha = 0.001$ slows the collapse, but entropy still converges to approximately 0.06, resulting in a performance drop on the benchmark. In contrast, increasing $\alpha$ to 0.003 triggers an entropy explosion, which is accompanied by a substantial degradation in performance.

In contrast, both the clip higher strategy in DAPO and our CE-GPPO method effectively suppress entropy collapse, maintaining entropy at a higher level while enabling steady improvements in model performance during training. Notably, DAPO exhibits an entropy rebound in the later stage (starting around step 300), which further suggests a potential issue of over-exploration. Conversely, CE-GPPO shows a slow but stable decline and achieves superior performance on the AIME25 benchmark. This indicates that CE-GPPO, when equipped with well-tuned $\beta_1$ and $\beta_2$, better balances exploration and exploitation.

\subsection{The Role of Entropy at Different Stages of Training}
\label{the_role_entropy_stage}
\begin{figure}[t]
    \centering
    \begin{subfigure}{0.48\linewidth}
        \centering
        \includegraphics[width=\linewidth]{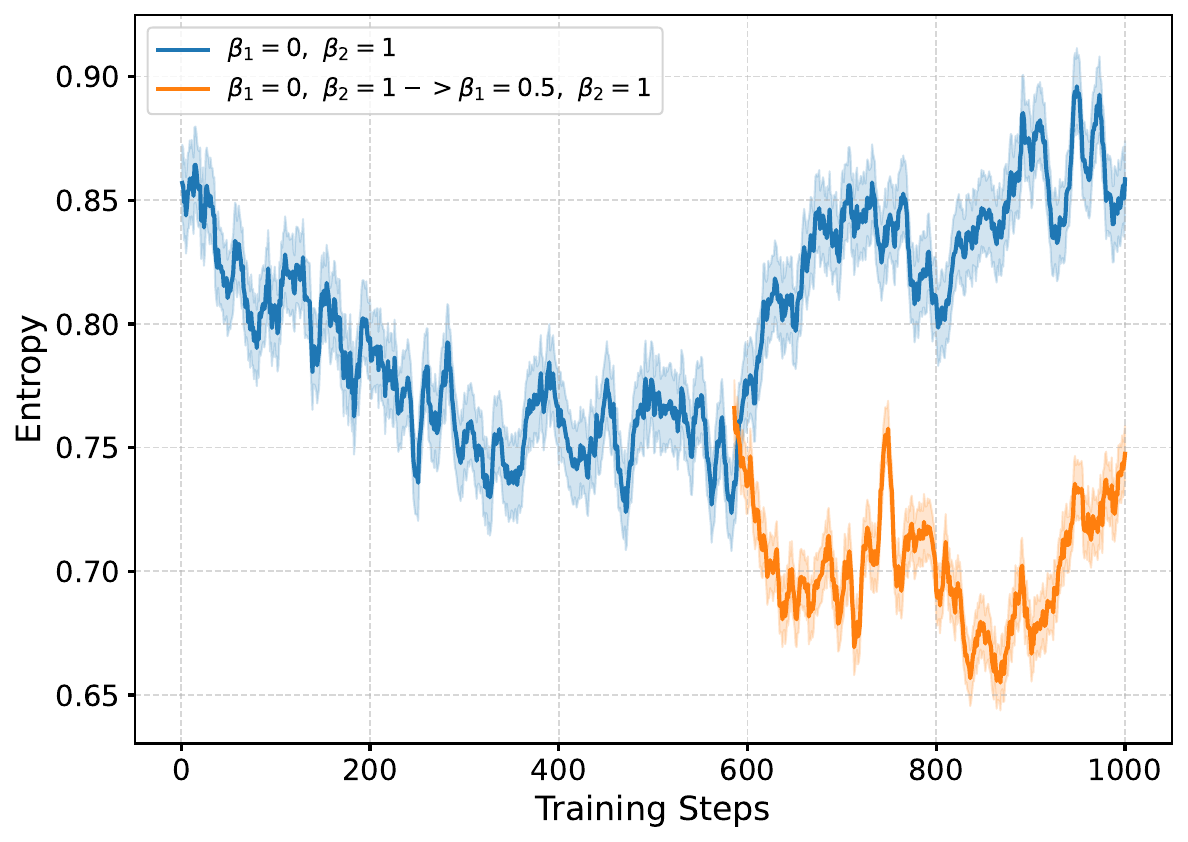}
        \caption{Entropy Dynamics}
        \label{fig:entropy_3}
    \end{subfigure}
    \hfill
    \begin{subfigure}{0.48\linewidth}
        \centering
        \includegraphics[width=\linewidth]{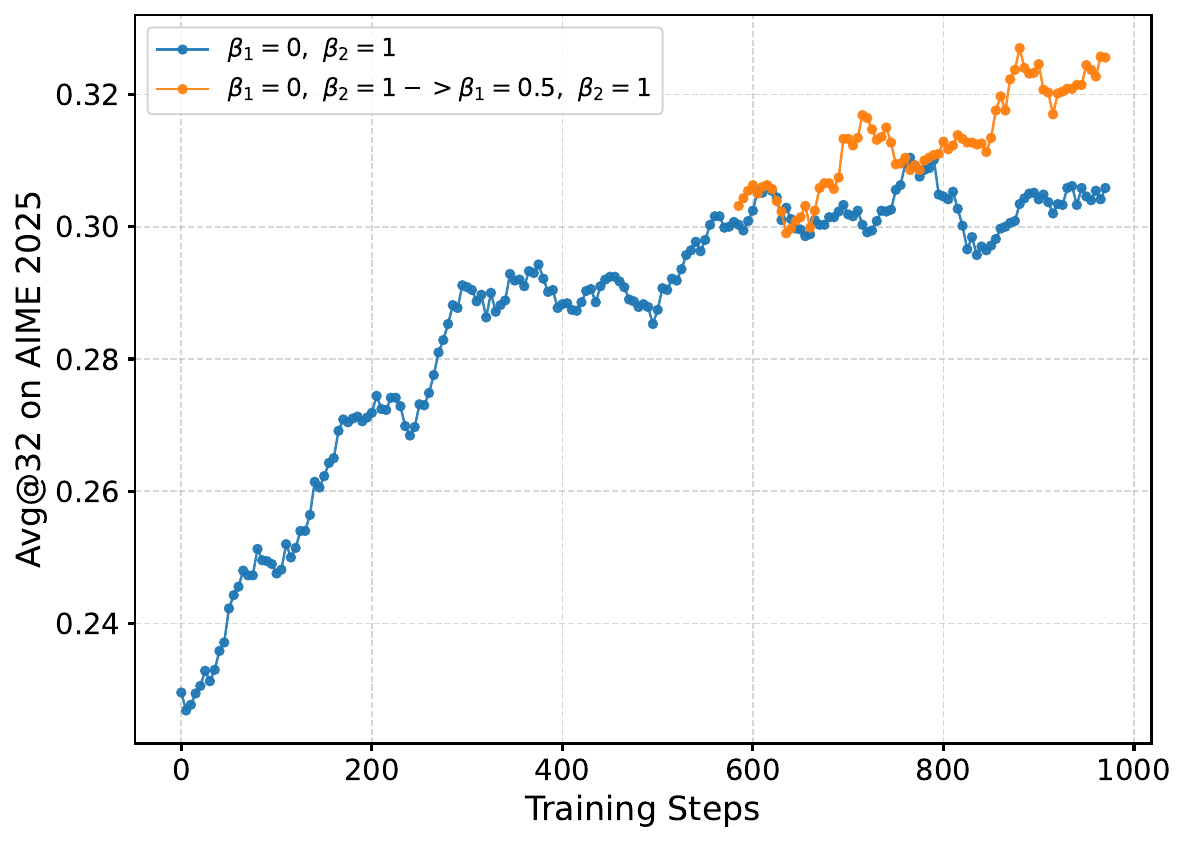}
        \caption{Benchmark Accuracy}
        \label{fig:entropy_4}
    \end{subfigure}
    \caption{Entropy dynamics and benchmark accuracy under different $\beta_1/\beta_2$ configurations. For $\beta_1=0/\beta_2=1$, the setting is maintained consistently across 0–1000 steps. For $\beta_1=0/\beta_2=1 \rightarrow \beta_1=0.5/\beta_2=1$ configuration, the transition occurs at step 585.}
    \label{fig:entropy_3_4}
\end{figure}

We further investigate the role of entropy at different stages of training. The results show that maintaining high and stable entropy in the early stage facilitates exploration, while in the later stages, a gradual convergence of entropy within a reasonable range helps stabilize performance and consolidate the learned knowledge. For example, in Figure ~\ref{fig:entropy_3_4}, under the setting $\beta_1=0,\beta_2=1$, entropy shows a slight upward trend in the late stage. By switching from $\beta_1=0,\beta_2=1$ to $\beta_1=0.5,\beta_2=1$ in the middle stage, entropy stabilizes and decreases in the later stage, which brings further performance improvements. This finding is consistent with the conclusion of Kimi K2 \citep{DBLP:journals/corr/abs-2507-20534} that exploration should be encouraged in the early stage and exploitation should be emphasized later. Our method is able to regulate the dynamics of entropy to achieve such a balance between exploration and exploitation, unlocking greater model performance.

\section{Conclusion}
In this paper, we investigate the intrinsic mechanisms that drive entropy dynamics in reinforcement learning for LLMs and identify clipped low-probability tokens as critical factors in balancing exploration and exploitation. Based on this understanding, we propose CE-GPPO, a novel algorithm that incorporates gradient signals from out-of-clip tokens in a controlled and theoretically grounded manner. By introducing a stop-gradient operation and tunable scaling, CE-GPPO effectively preserves training stability while enabling fine-grained control of policy entropy. Extensive experiments demonstrate that CE-GPPO prevents entropy collapse, avoids excessive exploration, and achieves superior performance compared to strong baselines including GRPO, DAPO, CISPO, and GSPO. 

\section*{Limitations}


Since each model has different parameters, its entropy pattern also varies; for example, some models exhibit high entropy at the beginning of RL training, while others start with low entropy. Although CE-GPPO demonstrates robustness to hyperparameters, we find that $\beta_1=0.5, \beta_2=1$ serves as a generally effective setting, achieving good performance across different models. For these diverse models, however, achieving optimal results still requires a certain degree of hyperparameter tuning, which we leave as future work.

\bibliography{custom}

\clearpage

\appendix
\section{Appendix}

\subsection{Experimental Setup for Other Baseline RL Methods}
\label{appendix-1}

\paragraph{GRPO} is a reinforcement learning algorithm designed to fine-tune LLMs by optimizing policies through group-based comparisons. Following \citep{DBLP:journals/corr/abs-2402-03300}, we set the clipping parameter $\epsilon$ for the upper and lower bounds to $0.2$.

\paragraph{DAPO} builds upon and refines the GRPO framework, addressing key limitations such as entropy collapse and training instability. Following \citep{DBLP:journals/corr/abs-2402-03300}, the lower and upper clipping thresholds were set to $\epsilon_l=0.2$ and $\epsilon_h=0.28$, respectively. 

\paragraph{CISPO} lies in the direct application of a clipping mechanism to the Importance Sampling (IS) weights, as opposed to clipping the final policy update~\citep{Cui2025TheEM}. We set the clipping parameter $\epsilon$ for the upper and lower bounds to $0.2$. 

\paragraph{GSPO} is a reinforcement learning algorithm developed to enhance the training stability, efficiency, and scalability of LLMs. The core of the algorithm lies in its use of a sequence-level importance ratio \citep{DBLP:journals/corr/abs-2507-18071}. Following \citep{DBLP:journals/corr/abs-2507-18071}, the lower and upper clipping thresholds were set to $\epsilon_l=0.0003$ and $\epsilon_h=0.0004$.

\subsection{Proof of the Formula for Policy Entropy Change}
\label{appendix-2}

\subsubsection{Problem Setup and Assumptions}

Let $\pi_\theta(a \mid s)$ denote a stochastic policy parameterized by $\theta$, where $s$ represents the state and $a$ represents an action. The policy entropy at state $s$ is defined as:

\begin{small}
\begin{equation}
\mathcal{H}(\pi_\theta \mid s) = -\sum_{a \in \mathcal{A}} \pi_\theta(a \mid s) \log \pi_\theta(a \mid s)
\end{equation}
\end{small}

We make the following standard assumptions:

\begin{itemize}[leftmargin=*]
    \item The policy follows a tabular softmax parameterization:
    \begin{small}
    \begin{equation}
    \pi_\theta(a \mid s) = \frac{\exp(z_{s,a})}{\sum_{a' \in \mathcal{A}} \exp(z_{s,a'})}
    \end{equation}
    \end{small}
    where $z_{s,a}$ are the logits corresponding to state-action pairs.
    
    \item Policy updates are performed using the policy gradient theorem with a sufficiently small learning rate $\eta > 0$, such that first-order Taylor approximations remain valid.
    
    \item The advantage function $\hat{A}(s,a)$ is baseline-centered, satisfying:
    \begin{small}
    \begin{equation}
    \mathbb{E}_{a' \sim \pi^k}[\hat{A}(s, a')] = 0
    \end{equation}
    \end{small}
    This is a common practice in policy gradient methods where advantages are computed as $Q(s,a) - V(s)$.
\end{itemize}

Our goal is to derive an approximation for the entropy change between successive policy updates:

\begin{small}
\begin{equation}
\Delta\mathcal{H} = \mathcal{H}(\pi_\theta^{k+1} \mid s) - \mathcal{H}(\pi_\theta^k \mid s)
\end{equation}
\end{small}

\subsubsection{Derivation}

\paragraph{Step 1: First-Order Taylor Expansion of Entropy}

Treating the entropy as a function of the logits $z_{s,a}$ at a fixed state $s$, we employ a first-order Taylor expansion around the current parameters $z^k$:

\begin{small}
\begin{equation}
\label{eq:taylor-expansion}
\mathcal{H}(\pi^{k+1} \mid s) \approx \mathcal{H}(\pi^{k} \mid s) + \sum_{a \in \mathcal{A}} \frac{\partial \mathcal{H}(\pi^{k} \mid s)}{\partial z_{s,a}} (z^{k+1}_{s,a} - z^{k}_{s,a})
\end{equation}
\end{small}

The entropy change can therefore be approximated as:

\begin{small}
\begin{equation}
\label{eq:entropy-change-approx}
\Delta\mathcal{H} \approx \sum_{a \in \mathcal{A}} \frac{\partial \mathcal{H}(\pi^{k} \mid s)}{\partial z_{s,a}} (z^{k+1}_{s,a} - z^{k}_{s,a})
\end{equation}
\end{small}

\paragraph{Step 2: Gradient of Entropy with Respect to Logits}

We now compute the gradient term $\frac{\partial \mathcal{H}(\pi^{k} \mid s)}{\partial z_{s,a}}$. Starting from the entropy definition:

\begin{small}
\begin{equation}
\mathcal{H}(\pi^k \mid s) = -\sum_{a' \in \mathcal{A}} \pi^k(a' \mid s) \log \pi^k(a' \mid s)
\end{equation}
\end{small}

Following standard derivations for the softmax parameterization, we obtain:

\begin{small}
\begin{equation}
\begin{aligned}
\label{eq:entropy-gradient}
\frac{\partial \mathcal{H}(\pi^k \mid s)}{\partial z_{s,a}} &= -\sum_{a} \frac{\partial \pi^k(a \mid s)}{\partial z_{s, a^{\prime}}}(\log \pi^k(a \mid s)+1) \\
& =-\sum_{a} \pi^k(a \mid s)\left(\mathbf{1}_{a=a^{\prime}}-\pi^k\left(a^{\prime} \mid s\right)\right)\\
& \quad (\log \pi^k(a \mid s)+1) \\
&= - \pi^k(a^{\prime} \mid s)(\log \pi^k(a^{\prime} \mid s) + 1) \\
& \quad + \pi^k(a^{\prime} \mid s) \sum_{a} \pi(a \mid s) (\log \pi^k(a \mid s) + 1) \\
&= -\pi^k(a \mid s) \left( \log \pi^k(a \mid s) \right.\\
& \quad \left. - \mathbb{E}_{a' \sim \pi^k}[\log \pi^k(a' \mid s)] \right)
\end{aligned}
\end{equation}
\end{small}

\paragraph{Step 3: Policy Gradient Update for Logits}

Under the policy gradient theorem and our baseline-centered advantage assumption, the logit update simplifies to:

\begin{small}
\begin{equation}
\begin{aligned}
\label{eq:logit-update}
z^{k+1}_{s,a} - z^{k}_{s,a} &= \eta \mathbb{E}_{a' \sim \pi_\theta(\cdot | s)} \left[ \frac{\partial \log \pi_\theta(a' | s)}{\partial z_{s,a}} \hat{A}(s, a') \right] \\
&= \eta \mathbb{E}_{a' \sim \pi_\theta} \left[ (\mathbf{1}_{a'=a} - \pi_\theta(a | s)) \hat{A}(s, a') \right] \\
&=  \eta \sum_{a'} \pi_\theta(a' | s)(\mathbf{1}_{a'=a} - \pi_\theta(a | s)) \hat{A}(s, a') \\
&= \eta \pi_\theta(a | s) \hat{A}(s, a) - \pi_\theta(a | s) \sum_{a'} \pi_\theta(a' | s) \hat{A}(s, a') \\
&= \eta \pi_\theta(a | s) \left( \hat{A}(s, a) - \mathbb{E}_{a' \sim \pi_\theta} [\hat{A}(s, a')] \right) \\
&= \eta \pi^k(a \mid s) \hat{A}(s, a)
\end{aligned}
\end{equation}
\end{small}

This follows directly from the policy gradient theorem and the baseline-centered advantage assumption $\mathbb{E}_{a' \sim \pi^k}[\hat{A}(s, a')] = 0$.

\paragraph{Step 4: Combining the Results}

Substituting Equations~\ref{eq:entropy-gradient} and~\ref{eq:logit-update} into the Taylor expansion from Equation~\ref{eq:entropy-change-approx}:

\begin{small}
\begin{equation}
\begin{aligned}
\Delta\mathcal{H} &\approx \sum_{a \in \mathcal{A}} \frac{\partial \mathcal{H}(\pi^{k} \mid s)}{\partial z_{s,a}} (z^{k+1}_{s,a} - z^{k}_{s,a}) \\
&= \sum_{a \in \mathcal{A}} \left[ -\pi^k(a \mid s) \left( \log \pi^k(a \mid s) - \right. \right. \\
& \left. \left. \mathbb{E}_{a' \sim \pi^k}[\log \pi^k(a' \mid s)] \right) \right] (z^{k+1}_{s,a} - z^{k}_{s,a}) \\
&= -\mathbb{E}_{a \sim \pi^k} \left[ \left( \log \pi^k(a \mid s) - \right. \right. \\
& \left. \left. \mathbb{E}_{a' \sim \pi^k}[\log \pi^k(a' \mid s)] \right) (z^{k+1}_{s,a} - z^{k}_{s,a}) \right] \\
&= -\left(\mathbb{E}_{a \sim \pi^{k}}\left[\log \pi^{k}(a \mid s)\left(z_{s, a}^{k+1}-z_{s, a}^{k}\right)\right] - \right. \\
& \left. \mathbb{E}_{a^{\prime} \sim \pi^{k}}\left[\log \pi^{k}\left(a^{\prime} \mid s\right)\right] \mathbb{E}_{a \sim \pi^{k}}\left[z_{s, a}^{k+1}-z_{s, a}^{k}\right]\right)
\end{aligned}
\end{equation}
\end{small}

This is precisely the definition of covariance:

\begin{small}
\begin{equation}
\begin{aligned}
\text{Cov}(X,Y) = \mathbb{E}[XY] - \mathbb{E}[X]\mathbb{E}[Y]
\end{aligned}
\end{equation}
\end{small}
where \begin{small}$X = \log \pi^k(a \mid s)$, $Y = z_{s,a}^{k+1} - z_{s,a}^k$\end{small}. Therefore:

\begin{small}
\begin{equation}
\begin{aligned}
\Delta\mathcal{H} \approx -\text{Cov}_{a\sim\pi^k}\left(\log \pi^k(a \mid s), z_{s,a}^{k+1} - z_{s,a}^k\right)
\end{aligned}
\end{equation}
\end{small}

Then we substitute the result obtained in step 3 to arrive at the following equation.

\begin{small}
\begin{equation}
\begin{aligned}
\Delta \mathcal{H} &\approx-\operatorname{Cov}_{a \sim \pi^{k}}\left(\log \pi^{k}(a \mid s), \eta \pi^{k}(a \mid s) \hat{A}(s, a)\right) \\
&= -\eta\operatorname{Cov}_{a \sim \pi^{k}}\left(\log \pi^{k}(a \mid s), \pi^{k}(a \mid s) \hat{A}(s, a)\right)
\end{aligned}
\end{equation}
\end{small}

This completes the derivation of the desired formula.

\subsection{Proof of the Gradient of CE-GPPO Objective}
\label{appendix:ce-gppo-gradient}

In this section, we derive the gradient of the proposed CE-GPPO loss function. Recall that the objective is defined as

\begin{small}
\begin{equation}
\begin{aligned}
\mathcal{J}_{\text{CE-GPPO}}(\theta) 
&= \mathbb{E} 
   \left[\frac{1}{\sum_{i=1}^G |y_i|} 
   \sum_{i=1}^G \sum_{t=1}^{|y_i|} 
   \ell^{(i)}\right],
\\[3pt]
\text{where} \quad
\ell^{(i)} &=
\begin{cases}
\beta_1 \cdot \, \dfrac{1-\epsilon}{\operatorname{sg}(\delta)} 
         \, \delta \, \cdot \hat{A}_{i,t}, 
& \text{if } \delta < 1-\epsilon \text{ and } \hat{A}_{i,t} < 0, \\[8pt]
\beta_2 \cdot \, \dfrac{1+\epsilon}{\operatorname{sg}(\delta)} 
         \, \delta \, \cdot \hat{A}_{i,t}, 
& \text{if } \delta > 1+\epsilon \text{ and } \hat{A}_{i,t} > 0, \\[8pt] 
\delta \cdot \, \hat{A}_{i,t}, 
& \text{otherwise}.
\end{cases}
\end{aligned}
\end{equation}
\end{small}

Here, $\delta$ is the importance sampling ratio between the updated and reference policies:

\begin{small}
\begin{equation}
\delta = \frac{\pi_\theta(y_{i,t} \mid y_{<t}, x)}{\pi_{\theta_{\text{old}}}(y_{i,t} \mid y_{<t}, x)}
\end{equation}
\end{small}

Note that $\operatorname{sg}(\cdot)$ denotes the stop-gradient operator, ensuring that the scaling terms $(1-\epsilon)/\operatorname{sg}(\delta)$ and $(1+\epsilon)/\operatorname{sg}(\delta)$ do not propagate gradients through $\delta$.

Taking the gradient of $\mathcal{J}_{\text{CE-GPPO}}(\theta)$, we obtain

\begin{small}
\begin{equation}
\begin{aligned}
\nabla_\theta \mathcal{J}_{\text{CE-GPPO}}(\theta) 
&= \mathbb{E} \left[
\frac{1}{\sum_{i=1}^G |y_i|} 
\sum_{i=1}^G \sum_{t=1}^{|y_i|}
\nabla_\theta \ell^{(i)}
\right]
\end{aligned}
\end{equation}
\end{small}

We now consider the three cases in the definition of $\ell^{(i)}$:

\paragraph{Case 1: $\delta < 1-\epsilon$ and $\hat{A}_{i,t} < 0$.}  

\begin{small}
\begin{equation}
\ell^{(i)} = 
\beta_1 \cdot \frac{1-\epsilon}{\operatorname{sg}(\delta)} \cdot \delta \cdot \hat{A}_{i,t}
\end{equation}
\end{small}

Since $\operatorname{sg}(\delta)$ is treated as a constant, the gradient only flows through $\delta$:

\begin{small}
\begin{equation}
\begin{aligned}
\nabla_\theta \ell^{(i)} 
&= \beta_1 \frac{(1-\epsilon)}{\frac{\pi_\theta(y_{i,t} \mid y_{<t}, x)}{\pi_{\theta_{\text{old}}}(y_{i,t} \mid y_{<t}, x)}} \cdot \frac{\nabla_\theta\pi_\theta(y_{i,t} \mid y_{<t}, x)}{\pi_{\theta_{\text{old}}}(y_{i,t} \mid y_{<t}, x)} \cdot \hat{A}_{i,t} \\
&= \beta_1 \frac{(1-\epsilon)}{\frac{\pi_\theta(y_{i,t} \mid y_{<t}, x)}{\pi_{\theta_{\text{old}}}(y_{i,t} \mid y_{<t}, x)}} \\
& \quad \cdot \frac{\pi_\theta(y_{i,t} \mid y_{<t}, x) \cdot \nabla_\theta \log \pi_\theta(y_{i,t} \mid y_{<t}, x)}{\pi_{\theta_{\text{old}}}(y_{i,t} \mid y_{<t}, x)} \cdot \hat{A}_{i,t} \\
&= \beta_1 (1-\epsilon) \cdot \hat{A}_{i,t} \cdot \nabla_\theta \log \pi_\theta(y_{i,t} \mid y_{<t}, x)
\end{aligned}
\end{equation}
\end{small}

\paragraph{Case 2: $\delta > 1+\epsilon$ and $\hat{A}_{i,t} > 0$.}  

\begin{small}
\begin{equation}
\ell^{(i)} = 
\beta_2 \cdot \frac{1+\epsilon}{\operatorname{sg}(\delta)} \cdot \delta \cdot \hat{A}_{i,t}
\end{equation}
\end{small}

Similarly,

\begin{small}
\begin{equation}
\nabla_\theta \ell^{(i)} 
= \beta_2 (1+\epsilon) \cdot \hat{A}_{i,t} \cdot \nabla_\theta \log \pi_\theta(y_{i,t} \mid y_{<t}, x)
\end{equation}
\end{small}

\paragraph{Case 3: Otherwise.}  

\begin{small}
\begin{equation}
\ell^{(i)} = \delta \cdot \hat{A}_{i,t}
\end{equation}
\end{small}

Thus,

\begin{small}
\begin{equation}
\nabla_\theta \ell^{(i)} 
= \delta \cdot \hat{A}_{i,t} \cdot \nabla_\theta \log \pi_\theta(y_{i,t} \mid y_{<t}, x)
\end{equation}
\end{small}

Combining the three cases, we may summarize the gradient as:

\begin{small}
\begin{equation}
\begin{aligned}
\nabla_\theta \mathcal{J}_{\text{CE-GPPO}}(\theta) &= 
\mathbb{E} \left[
\frac{1}{\sum_{i=1}^G |y_i|} 
\sum_{i=1}^G \sum_{t=1}^{|y_i|}
\mathcal{F}_{i,t}(\theta) \right. \\
& \quad \left. \cdot \nabla_\theta \log \pi_\theta(y_{i,t} \mid y_{<t}, x) \cdot \hat{A}_{i,t}
\right],
\end{aligned}
\end{equation}
\end{small}

where the weighting factor $\mathcal{F}_{i,t}(\theta)$ is defined as

\begin{small}
\begin{equation}
\mathcal{F}_{i,t}(\theta) =
\begin{cases}
\beta_1 \cdot (1-\epsilon), & \text{if } \delta < 1-\epsilon \text{ and } \hat{A}_{i,t} < 0, \\[0.5em]
\beta_2 \cdot (1+\epsilon), & \text{if } \delta > 1+\epsilon \text{ and } \hat{A}_{i,t} > 0, \\[0.5em]
\delta, & \text{otherwise}.
\end{cases}
\end{equation}
\end{small}

This completes the proof.

\end{document}